%% file: main.tex
\documentclass{article}
\usepackage{ijcai24}

\usepackage{times}
\usepackage{soul}
\usepackage{url}
\usepackage[hidelinks]{hyperref}
\usepackage[utf8]{inputenc}
\usepackage[small]{caption}
\usepackage{graphicx}
\usepackage{amsmath}
\usepackage{booktabs}
\usepackage{algorithm}
\usepackage{algorithmic}
\usepackage[switch]{lineno}

\usepackage{subcaption}
\newcommand{\mydarkblue}[1]{\textcolor[rgb]{0.2,0.5,0.8}{ #1}}
\newcommand{\cmark}{\ding{51}}%
\newcommand{\xmark}{\ding{55}}%
\usepackage{pifont}
\usepackage[misc]{ifsym}

\usepackage{epsfig}
\usepackage{graphicx}
\usepackage{amsmath}
\usepackage{amssymb}
\usepackage{multirow} 
\usepackage{booktabs}
\usepackage{colortbl}
\usepackage[table,xcdraw]{xcolor} 
\usepackage{xspace} 
\usepackage{hhline} 
\usepackage[capitalize]{cleveref}
\input{math_commands}

\newcommand{\upscore}[1]{\footnotesize{\mydarkred{$\blacktriangledown$ #1\%}}}
\newcommand{\mydarkred}[1]{\textcolor[rgb]{0.8,0.0,0.0}{ #1}}
\newcommand{\tabincell}[2]{\begin{tabular}{@{}#1@{}}#2\end{tabular}}

\title{ELF-UA: Efficient Label-Free User Adaptation in Gaze Estimation}

\author{
Yong Wu$^1$
\and
Yang Wang$^2$
\and
Sanqing Qu$^1$
\and
Zhijun Li$^1$
\And
Guang Chen$^{1(\textrm{\Letter})}$\\
\affiliations
$^1$Tongji University\\
$^2$Concordia University\\
\emails
\{yongwu, 2011444, guangchen\}@tongji.edu.cn,
yang.wang@concordia.ca, zjli@ieee.org
}

\begin{document}

\maketitle
\input{sec/0_abstract}
\input{sec/1_intro}
\input{sec/2_related}
\input{sec/3_method}

\input{sec/4_experiment}

\input{sec/5_conclusion}

\bibliographystyle{named}
\bibliography{ijcai24}

\input{sec/6_appendix}

\end{document}

%% file: math_commands.tex
\usepackage{amsmath,amsfonts,bm}

\def\eqref#1{equation~\ref{#1}}

\def\1{\bm{1}}

\DeclareMathAlphabet{\mathsfit}{\encodingdefault}{\sfdefault}{m}{sl}
\SetMathAlphabet{\mathsfit}{bold}{\encodingdefault}{\sfdefault}{bx}{n}



%% file: sec/0_abstract.tex
\begin{abstract}
   We consider the problem of user-adaptive 3D gaze estimation. The performance of person-independent gaze estimation is limited due to interpersonal anatomical differences. Our goal is to provide a personalized gaze estimation model specifically adapted to a target user. Previous work on user-adaptive gaze estimation requires some labeled images of the target person data to fine-tune the model at test time. However, this can be unrealistic in real-world applications, since it is cumbersome for an end-user to provide labeled images. In addition, previous work requires the training data to have both gaze labels and person IDs. This data requirement makes it infeasible to use some of the available data. To tackle these challenges, this paper proposes a new problem called efficient label-free user adaptation in gaze estimation. Our model only needs a few unlabeled images of a target user for the model adaptation. During offline training, we have some labeled source data without person IDs and some unlabeled person-specific data. Our proposed method uses a meta-learning approach to learn how to adapt to a new user with only a few unlabeled images. Our key technical innovation is to use a generalization bound from domain adaptation to define the loss function in meta-learning, so that our method can effectively make use of both the labeled source data and the unlabeled person-specific data during training. Extensive experiments validate the effectiveness of our method on several challenging benchmarks.
\end{abstract}

%% file: sec/1_intro.tex
\section{Introduction}

\begin{figure}[ht]

	\begin{center}
		\includegraphics[width=3.2in]{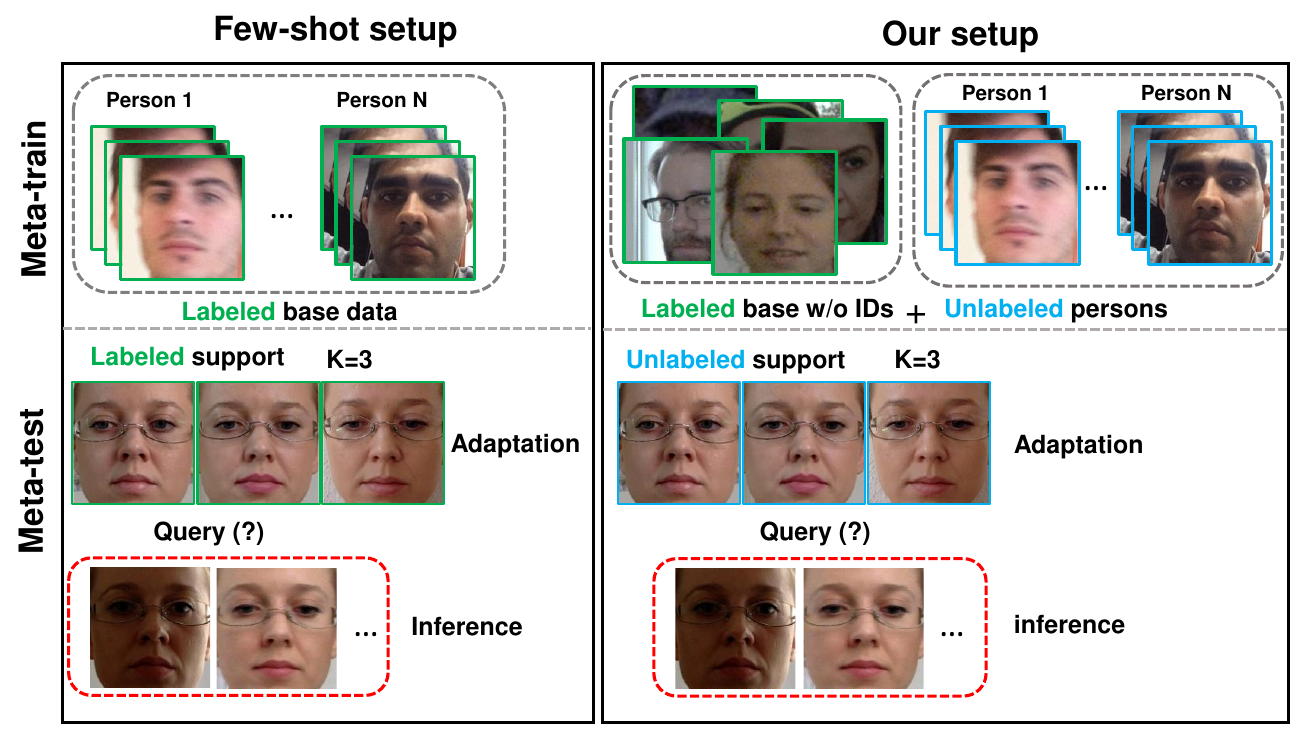}
	\end{center} 
	\caption{ { \bf Illustration of our problem setup.} (Left) In previous work on few-shot user adaptive gaze estimation~\protect\cite{park2019few}, the model is trained on the labeled base dataset with many persons (also called \emph{tasks}) during meta-training. During meta-testing, given a few labeled samples from a new person (known as \emph{support set}), the model is adapted to the new person. Then the model predicts other images (called \emph{query set}) from this person. (Right) In this work, we propose a new problem setting called label-free user adaptation in gaze estimation. During meta-training, we have a labeled source dataset without person IDs. We also have some unlabeled data with person IDs. During met-testing, we are provided very few \emph{unlabeled} images ($\le$5 ) from a new target person. Our goal is to get a gaze model specifically adapted to this target person. Note that our problem setup does not require any labeled images from the target person for adaptation. Our problem setting is closer to real-world scenarios.}
 \label{fig:problem}
	\vspace{-0.3cm}
\end{figure}
Gaze is a crucial non-verbal cue for understanding the internal states of humans. Gaze estimation from images has been widely used in various applications, such as saliency detection~\cite{2019wenguang}, human-computer interaction~\cite{2017Zhang1}, and virtual reality industry~\cite{prange2017speech,2020Chen,konrad2020gaze}. In this paper, we focus on predicting 3D gazes since it has more applications than 2D gazes.  Most existing gaze estimation approaches~\cite{2017Zhang1,2017Zhang4,2016Krafka} use standard supervised learning to learn a person-independent model. As observed in \cite{park2019few}, the performance of person-independent gaze estimation is limited due to inter-personal anatomical differences. In order to achieve the high quality required by many applications, it is much more desirable to personalize the model to each user. Unfortunately, it is unrealistic to train a personalized gaze estimation for each user using standard supervised learning since it is impossible to collect enough training data for each user. Some recent work~\cite{park2019few} proposes a setting called few-shot adaptive gaze estimation. It uses a meta-learning approach to learn a model during offline training (called ``meta-training'' in meta-learning). During testing (called ``meta-testing'' in meta-learning), the model is adapted to a new user with a few labeled images of this user.

Compared with standard supervised learning, the few-shot adaptive gaze estimation setting in \cite{park2019few} represents a significant advancement toward real-world use. However, this setting still has limitations that prevent practical deployment. First, during meta-testing, it requires a few \emph{labeled} images of the target user in order to adapt the model to a new user. In practice, it is unrealistic to ask users to provide annotations in real-world applications. This is especially for 3D gaze, since the 3D gaze annotation may require specialized hardware (e.g. Tobii eye tracker) inaccessible to regular users. In contrast, it is much easier for the end user to simply capture a small number (e.g. 1 to 5) of unlabeled images. It is much more desirable if we could adapt the model using only a few unlabeled images of the target user. Second, during meta-training, the work in \cite{park2019few} requires the training data to be collected from different persons, where each person has multiple images. Each training image has to be fully annotated with both the gaze label and the person ID. The person ID is needed to construct the tasks in meta-learning, where each task corresponds to a user. This puts too much constraint on the training data. On one hand, some existing gaze estimation datasets~\cite{gaze360_2019} do not come with person IDs. The method in \cite{park2019few} cannot make use of these datasets. For some large companies (e.g. Google, Amazon and Baidu), they are easy to collect many unlabeled face images without person IDs. However, the annotations are expensive. On the other hand, many 3D gaze datasets are collected in controlled environments~\cite{eth_2020,2014Funes} due to the difficulty of obtaining ground-truth 3D gaze labels. For training 3D gaze estimation models for real-world applications, we eventually have to collect new data to cover diverse real-world scenarios (e.g. users interacting with cell phones, tablets, smart TVs, etc). It can be difficult to collect labeled images since these images might be crowd-sourced from end users. Unlike other computer vision tasks (e.g. image classification) where ground-truth labels can be annotated by humans in a post hoc way, the 3D gaze label has to be obtained from specialized hardware (e.g. Tobii eye tracker) at the time when the image itself is taken. Due to the domain shift between existing gaze datasets and different users in real-world scenarios, it is not optimal to directly deploy a gaze model learned from existing labeled gaze datasets to a new user. In order to enable new applications of gaze estimation in the future, we need to come up with a way of making use of the diverse data even if they are unlabeled.

To address the aforementioned issues, we propose a new problem setting called efficient label-free user adaptation in gaze estimation. As shown in Fig.~\ref{fig:problem}, there are two key differences between our setting and the few-shot setting in \cite{park2019few}. First of all, the work in \cite{park2019few} requires a small number (e.g. 1 to 5) of \emph{labeled} images to adapt to a new user during meta-testing. In contrast, we only need a small number (e.g. 1 to 5) of \emph{unlabeled} images for the model adaptation. Second, all training data in \cite{park2019few} have to be annotated with both gaze labels and person IDs. In contrast, our method only requires a source dataset annotated with gaze labels (without person ID) and some unlabeled person-specific dataset with person ID (but without gaze labels). In practice, the source data can be any existing gaze dataset. The unlabeled person-specific data can be easily obtained by crowd-sourcing from end users without requiring specialized hardware or manual annotation. Table~\ref{tab:diff} highlights the differences between our problem setting and other related problem settings, including unsupervised domain adaptation (UDA)~\cite{cai2020generalizing,liu2021generalizing}, source-free domain adaptation (source-free DA)~\cite{pmlrliang20a} and few-shot learning (FSL)~\cite{park2019few}.
\input{tabs/tabdiff}

Our proposed method is based on meta-learning, in particular model-agnostic meta-learning (MAML) \cite{2017_finn_MAML}. Our network architecture consists of a main branch for the gaze prediction and an auxiliary branch for a self-supervised task with a shared backbone. Each user is considered a task in MAML. During meta-testing, we adapt the model to a target user using only a few unlabeled images by optimizing the loss function of the self-supervised auxiliary task. During meta-training, the model is trained to facilitate effective fast adaptation to each user using a bi-level optimization. The main challenge is that since the person-specific images during meta-training are unlabeled, we cannot easily define a loss for the outer loop of MAML as in \cite{park2019few,2017_finn_MAML}. The key innovation of our work is to introduce a surrogate loss based on a theoretical result from domain adaptation for the outer loop in MAML. 

\noindent{{\bf Summary of Contributions:}} The contributions of this work are manifold. First, we present a new problem called efficient label-free user adaptation in
gaze estimation. Different from the standard supervised learning~\cite{2017Zhang1} or adaptation adaptation setup~\cite{park2019few,2019Yu1}, our problem formulation for training uses some labeled source data without person IDs and some unlabeled person-specific data, which learns a better person-specific gaze model. For an unseen person, the trained model only uses unlabeled samples to adapt the gaze model for this person.
Second, we propose a novel approach to address this problem, which uses generalization bound from domain adaptation to define the loss function in meta-learning. Our model can effectively adapt to a new person given the a very few unlabeled images from that person at test time. Finally, we conduct an extensive evaluation of the proposed approach on several challenging benchmark datasets. Our approach significantly outperforms other alternatives and achieves competitive performance compared with other state-of-the-art methods.

%% file: tabs/tabdiff.tex
\begin{table*}[htbp]
\centering
\caption{ {\bf Comparison of different problem settings in gaze estimation.} 
If a method only requires a few gradient updates with a small number of data, we consider it to be efficient. We also indicate whether the adapted model is dataset-specific or person-specific. For example, UDA~\protect\cite{liu2021generalizing} requires access to the source data during adaptation. Source-free DA~\protect\cite{qu2022bmd} removes this assumption. Both methods~\protect\cite{liu2021generalizing,qu2022bmd} require enough unlabeled target data. Both of them produce a model adapted to a particular dataset (not a specific user) in the end. FSL~\protect\cite{park2019few} can efficiently adapt to a target user with only a few gradient updates. But it requires a few labeled examples of the target user for adaption. Our proposed setting only needs a few unlabeled images of a user for adaptation. Our approach is efficient since it only needs a few gradient updates for adaptation. In addition, our approach can make sure of both labeled and unlabeled data during training, as long as the unlabeled data come with person IDs.}
		\label{tab:diff}
		
		  \small
			\resizebox{1\linewidth}{!}{
			\begin{tabular}{c|cc|cccc|c}
			\hline	
                \multirow{2}{*}{Method}& \multicolumn{2}{c|}{Training} & \multicolumn{4}{c|}{Adaptation} & {Inference}\\
                \cline{2-8}
                 & Source data & Source labels & Source data & Target data & Target labels & Efficient & Model\\ 
		      \hline
				UDA~\cite{liu2021generalizing} & \mydarkblue{$\blacksquare\blacksquare\blacksquare\blacksquare$} & \mydarkblue{$\blacksquare\blacksquare\blacksquare\blacksquare$}
                & \cmark 
                &\mydarkblue{$\blacksquare\blacksquare\blacksquare\blacksquare$}
                & \xmark 
                & No & Dataset-specfic\\
                Source-free DA~\cite{qu2022bmd} &
                \mydarkblue{$\blacksquare\blacksquare\blacksquare\blacksquare$} & \mydarkblue{$\blacksquare\blacksquare\blacksquare\blacksquare$} &
                \xmark 
                & \mydarkblue{$\blacksquare\blacksquare\blacksquare\blacksquare$}
                & \xmark & No & Dataset-specific\\

                FSL~\cite{park2019few} &
                \mydarkblue{$\blacksquare\blacksquare\blacksquare\blacksquare$} & \mydarkblue{$\blacksquare\blacksquare\blacksquare\blacksquare$} &
                \xmark & 
                \mydarkblue{$\blacksquare$} &
                \cmark & Yes & Person-specific\\
                
                Ours &
                \mydarkblue{$\blacksquare\blacksquare\blacksquare\blacksquare$} & \mydarkblue{$\blacksquare\blacksquare$} &
                \xmark & 
                \mydarkblue{$\blacksquare$} &
                \xmark & Yes & Person-specific\\
                \hline
		\end{tabular} 
            }

\end{table*}

%% file: sec/2_related.tex
\section{Related Work}
In this section, we review prior work in four main lines of
research most relevant to our work, namely gaze estimation, domain adaptation, meta-learning and self-supervised learning.

\noindent{\bf Gaze Estimation.}
Most previous work in gaze estimation users supervised learning. These methods require large-scale labeled data. However, it is hard to obtain a large amount of labeled data, especially for 3D gaze which requires specified hardware (e.g. Tobii eye tracker) to obtain the ground-truth label. In addition , due to the different anatomical structures (e.g. eye shape, visual axis) between different persons, the performance of supervised gaze estimation models is limited when testing on unseen persons. Some recent work~\cite{lu2014pami,park2019few,2019yu_cvprw,2019Yu1,liu2024testtime} propose to build personalized gaze estimation specifically for each target user. For example, fine-tuning a pretrained model ~\cite{He_2019_ICCV_Workshops,2019Yu1,park2019few} can achieve very high accuracy for a specific person, but it usually requires annotated 
data from this person during testing. This requirement leads to difficulties in real-world 
applications. In our work, we aim to address this issue with label-free user adaptation where the adaptation only requires a few unlabeled images of the target user.

\noindent{\bf Domain Adaptation.} 
Domain adaptation (DA) has been used to address the domain shift between two domains by transferring the knowledge from the source domain to the target domain ~\cite{liu2021generalizing,cai2020generalizing,pmlrliang20a,qu2022bmd,cai2019learning,xu2020joint,chi2024adapting,wu2024testtime,peng2024map}. It can minimize statistical discrepancy across domains. For gaze estimation, there is often a domain shift between datasets due to differences in terms of image capture devices and environments. There is also domain shift between different users due to inter-personal anatomical difference. Adversarial~\cite{cui2020gradually,long2018conditional} learning is applied to explore unaligned feature space between source dataset and target dataset. However it has access to target samples in advance of inference. Unsupervised domain adaptation (UDA)~\cite{liu2021generalizing} is another direction of research to alleviate domain shift, which has no access to target labels during training. It also needs target samples. Few-shot learning (FSL)~\cite{park2019few,2019Yu1} is also adopted to improve performance when there is indistinguishable between different domains. Nevertheless, FSL needs a large number of labeled person-specific data to train the model during meta-training, then update the model with few labeled target samples during meta-testing. It is unrealistic in a real-world, since the acquisition of labeled data 
 during testing is very difficult. In contrast, our method only requires a labeled source dataset without person ID and unlabeled person-specific dataset to train a personalized gaze model. The source dataset can be any existing gaze dataset or collected from web. Further, our method uses a few unlabeled target samples to update the trained model for each unseen person at test time.

\noindent{\bf Meta-Learning.} Recently, 
meta-learning~\cite{shen21_aaai,2017_finn_MAML,2018_reation_net,chen2019closerfewshot,Lee_2019_CVPR,2016_matchingnet,2017_protonet,wu2022few,zhong2023metadmoe,Chi_2022_CVPR}
has been explored for quick adaptation to new categories. Meta-learning can learn transferable information for various tasks, which is critical in few-shot learning. More specifically, the existing meta-learning methods can be categorized as the metric-based~\cite{2017_protonet,2018_reation_net}, model-based~\cite{metanetwork} and optimization-based~\cite{2017_finn_MAML,2018firstorder}. Our proposed approach builds upon one of the most popular meta-learning algorithms, namely model agnostic
meta-learning (MAML)~\cite{2017_finn_MAML}. MAML involves a bi-level optimization, including inner loop and outer loop. The 
goal of MAML is to learn a model initialization such that it can be quickly adapted 
to any new task with a few labeled data. Different from standard MAML, our proposed approach does not require labels of tasks during meta-training, and does not require any labeled data for adaptation. 

\noindent{\bf Self-Supervised Learning.}
Self-supervised learning~\cite{Chi_2021_CVPR,carlucci2019domain,gidarisunsupervised,tang2021self,liang2022self,an2021conditional} is often used to learn feature representations by solving a pretext task, where the label of the pretext task can be obtained from an image itself. In computer vision, many self-supervised pretext tasks have been proposed, such as rotation degrees~\cite{gidarisunsupervised} or shuffling order of patch-shuffled images~\cite{carlucci2019domain}. Recently, there have some advanced self-supervised learning approaches using contrastive learning. Typically, self-supervised approaches learn generic feature representations which mitigate over-fitting to domain-specific biases. Carlucci~\emph{et al.}~\cite{carlucci2019domain} propose to solve a jigsaw puzzle as the auxiliary task, 
where the image is cropped the shuffled image patches and the model is trained to learn the right orders. This helps the model to learn the concepts of spatial correlation while acting as a regularizer for the classification task. In this work, we use the jigsaw puzzle as a self-supervised task for model adaptation in gaze estimation. Since the whole face images provide rich gaze information~\cite{cheng2020coarse,wu2022} than eyes. Reorder the face image at the training allows the model to learn internal gaze information.

\section{Preliminaries}

In this section, we first describe our problem setting, then we introduce a generalization bound that is the foundation of many domain adaptation approaches. Our proposed method builds upon this generalization bound.

\subsection{Problem Setting}
During offline training (called ``meta-training'' using meta-learning terminology), we are given a labeled source dataset $\mathcal{S}=\{(x_i,y_i)\}_{i=1}^{N_s}$, where $x_i$ is a training image and $y_i$ is the corresponding gaze label. The instances of the source data are \emph{not} person specific. In other words, we do not know the person ID of each instance (e.g. dataset Gaze360~\cite{gaze360_2019}). It is possible for each instance in $\mathcal{S}$ to be of a unique person. The standard supervised gaze estimation usually trains a model $f_{\theta}(\cdot)$ with parameter $\theta$ directly using $\mathcal{S}$. This model is then used during testing for any user.

In addition to the source dataset $\mathcal{S}$, we also have access to an \emph{unlabeled person-specific} dataset $\mathcal{T}$. This person-specific dataset contains unlabeled images of $K$ persons, where each person has multiple images, i.e. $\mathcal{T}=\{\mathcal{D}_1, \mathcal{D}_2, ... \mathcal{D}_K\}$ where $\mathcal{D}_i=\{x_j\}_{j=1}^{N_i}$ is the subset of $N_i$ images corresponding to the $i$-th person. We assume $N_i$ is reasonably large In practice, this person-specific dataset can be fairly easily captured using smartphones or web cameras. We assume that there is a domain shift between the source dataset and each user, i.e. $P_{\mathcal{S}}\neq P_{\mathcal{D}_i}$. There is also a domain shift between two different users, i.e. $P_{\mathcal{D}_i}\neq P_{\mathcal{D}_j}$ for $i\neq j$. Due to this domain shift, a model learned from $\mathcal{S}$ will not work well for a particular person $\mathcal{D}_i$. Ideally, we would like to have a personalized model for each user.

During testing (called ``meta-testing'' using meta-learning terminology), we have a new user. For this user, we only have \emph{few-shot unlabeled} images $\mathcal{D}=\{x_j\}_{j=1}^{d}$ where $d$ is small (e.g. 1 to 5). Our goal is to adapt the model $\theta$ using $\mathcal{D}$ and obtain an adapted model $\theta'$. Then the adapted model $\theta'$ will be used for prediction for new images of this user. 

\noindent{\bf Remark.} In the real-world scenarios, many systems tuned to specific environment or person need to be calibrated. we can easily collect some images via web cameras (collecting many enough images are still challenging), but acquisition of labels is very difficult.  This leads to inferior performance even if the system is calibrated by unlabeled images. Motivated by this, we propose this new problem setting, which pans out for this problem. We can update the system tailored to a new environment or person with a very few samples. Existing methods in the literature cannot address this new problem setting. Standard few-shot setting in~\cite{park2019few,2019Yu1} needs labeled images of target persons for adaptation, while unsupervised domain adaptation or source-free domain methods~\cite{cai2020generalizing,cui2020gradually} require target person images to align the model during training.

\subsection{Generalization Bound in Domain Adaptation}
In this section, we briefly introduce a mathematical tool in analyzing generalization bound
for domain adaptation~\cite{mansour2009domain,ben2010theory}. Our proposed method builds upon this generalization bound.

Let $\mathcal{S}$ be a source domain and $\mathcal{T}$ be a target domain. There is a domain shift between these two domains. Given a machine learning model $f_{\theta}$ that maps an input $x$ to an output $y$, $L(f_{\theta}(x),y)$ be a loss function that measures the difference between the prediction $f_{\theta}(x)$ and the ground-truth $y$. 
Then the following bound holds:
\begin{equation} \label{eq:bound}
\textbf{Theorem 1.}~~\mathbb{E}_{\mathcal{T}}[L(f_\theta(x), y)]\leq  \mathbb{E}_{\mathcal{S}}[L(f_\theta(x), y)] + d(\mathcal{S}, \mathcal{T}),
 \qquad
\end{equation}
where $\mathbb{E}_{\mathcal{S}}[\cdot]$ (or $\mathbb{E}_{\mathcal{T}}[\cdot]$) denotes the expectation with respect to the distribution of the domain $\mathcal{S}$ (or $\mathcal{T}$). Here $d(\mathcal{S},\mathcal{T})$ is a measure that quantifies the difference between $\mathcal{S}$ and $\mathcal{T}$. In practice, $d(\cdot)$ is defined such that it does not require labels in the target domain.

\textcolor{blue}{Theorem 1} provides the theoretical foundation for many domain adaptation methods. In domain adaptation, we would ideally like to minimize the loss $\mathbb{E}_{\mathcal{T}}[L(f_\theta(x), y)]$ on the target domain. However, this loss is not well defined on the target domain since we only have unlabeled data in the target domain. Instead, domain adaptation methods usually optimize the upper bound on the right hand side of \textcolor{blue}{Theorem 1}. Maximum Mean Discrepancy (MMD)~\cite{long2017deep} is a popular method to measure the distance between different domains, and we can also use it to solve $d(\mathcal{S},\mathcal{T})$ in \textcolor{blue}{Theorem 1}. Note that in order to  use MMD for domain adaption, we need to know the target domain in advance, and the unlabeled data in the target domain should be large enough.

%% file: sec/3_method.tex
\section{Our Approach}

\begin{figure*}[ht!]
	\begin{center}
		\includegraphics[width= 1\textwidth]{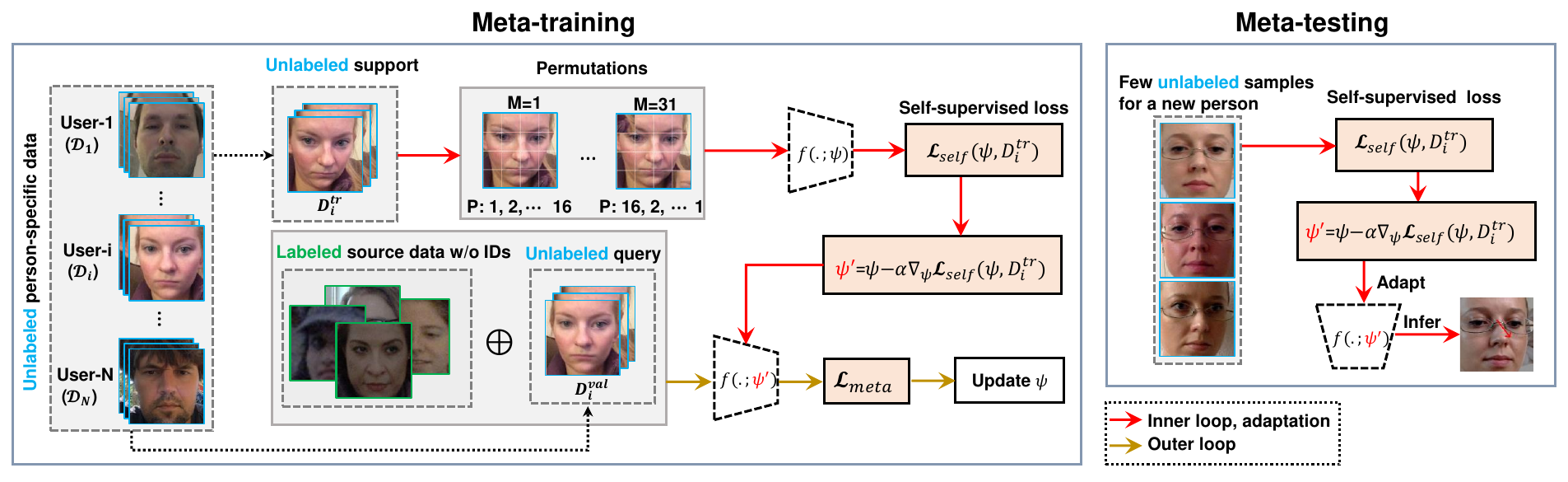}
	\end{center}
 \vspace{-0.5cm}
	\caption{{\bf Overview of our approach.} Our training data consist of a source dataset $\mathcal{S}$ annotated with gaze labels but without person IDs. We also have an unlabeled subject-specific dataset, where images are annotated with person IDs but without gaze labels. Each subject corresponds to a ``task'' in meta-learning. For each task $i$, we construct its support set $D^{tr}_i$ and its query set $D^{val}_i$, both of which consist of unlabeled images. In the inner loop of meta-training, we use a self-supervised auxiliary task (permutation prediction) defined on the support set $D^{tr}_i$ to update the model $\psi$ to a user-adapted model $\psi'$. Since the query set $D^{val}_i$ is unlabeled, we cannot directly compute a supervised loss of $\psi'$ on $D^{val}_i$. Instead, we use a domain adaptation loss defined on $\mathcal{S}$ and $D^{val}_i$ as an upper bound approximation to the supervised loss on $D^{val}_i$. This domain adaptation loss is used for the outer loop of meta-training. After meta-training, the model has learned to effectively adapt to a new subject using only a few unlabeled images. During meta-testing, we adapt the model to a new subject and use the adapted model for inference.} 
  \vspace{-0.3cm}
	\label{fig:meta}
\end{figure*}

We propose a novel meta-learning framework for learning self-supervised user-adaptive gaze estimation. Our proposed framework is based on the model-agnostic meta learning (MAML)~\cite{2017_finn_MAML}. The architecture overview is illustrated in Fig.~\ref{fig:meta}.
During meta-training, we learn to transfer useful knowledge from the labeled source dataset $\mathcal{S}$ and the unlabeled person-specific dataset $\mathcal{T}$ to achieve generalization on unseen persons. We consider each person in $\mathcal{T}$ as a task in meta-learning. For each task, we construct a task with a support set $D^{tr}$ and query set $D^{val}$. The support set $D^{tr}$ is used to update the model parameter $\psi$ to obtain a user-adapted model $\psi'$. The query set $D^{val}$ is used to measure the performance of the adapted model $\psi'$. 
The meta-training procedure involves bi-level optimization as shown in Fig.~\ref{fig:meta}, where the inner loop performs task-level optimization, 
while the outer loop performs a global model update via meta-objective.

\noindent{\bf Self-Supervised Inner Update.} In the inner loop of meta-training, we are given a sampled task (i.e. person) $i$. We use $D_i^{tr}$ to denote the support set of this task. Different from \cite{park2019few}, our support set $D_i^{tr}$ only contains a few unlabeled images of this person. Our goal is to update the model $\psi$ using $D_i^{tr}$ to obtain a user-adapted model $\psi'_i$. 

Inspired by \cite{TTT}, we add an additional head to our network for a self-supervision auxiliary task. Fig.~\ref{fig:meta} shows the illustration of the proposed model. The face images are decomposed to 16 patches with a $4\times4$ grid, which are randomly shuffled and then recomposed the face images. We define a set of $M$ patch permutations and assign an index to each of them. The original ordered and the shuffled images are passed to the network. The auxiliary task is to predict the permutation from the shuffled image. The face images enable to provide more gaze cues~\cite{cheng2020coarse,wu2022}. During the inner loop, the overall self-supervised loss on the support set is defined as:
\begin{equation}
\mathcal{L}_{self}(\psi; D_i^{tr}) = \sum_{x\in D_i^{tr}} 
\ell_{M}(f(z_k|x), M_k).\label{eq:loss_inner}
\end{equation}
Note that $z_k$ is the index for the input shuffled image, and $M_k$ is the index for the original order. $\ell_M$ is the standard cross-entropy loss for predicting the right order. During the inner loop, we do not need any labels for training the model, which is able to exploit the internal gaze information by solving the auxiliary self-supervised task. 

Our goal is to update the model parameters $\psi$ to obtain a person-adaptive model $\psi'_i$ specifically tuned to the task $\mathcal{D}_i$. We obtain $\psi'_i$ by taking a small number of gradient updates using Eq.~\ref{eq:loss_inner} as:
\begin{equation}
\psi'_i \leftarrow \psi-\alpha \nabla_{\psi}  \mathcal{L}_{self}(\psi; 
D_i^{tr}).
\label{eq:inner}
\end{equation}

\noindent{\bf Domain Adaptation Based Outer Update.} The adapted model $\psi'_i$ is specifically tuned to the $i$-th person. Intuitively, we would like $\psi'_i$ to perform well on other images of the same person. In previous applications of  MAML~\cite{2017_finn_MAML,park2019few}, one can uses the loss defined on the query set $D^{val}_{i}$ as this performance measure. However, since $D_i^{val}$ only contains unlabeled images in our setting, it is not trivial how to directly define a loss using $D^{val}_{i}$.

Our key insight is that we can approximate the supervised loss defined on $D^{val}_{i}$ by its upper bound provided by RHS of \textcolor{blue}{Theorem 1}.  We can measure the performance of $\psi'_i$ on $D_i^{val}$ of $\mathcal{D}$ as:
\begin{equation}
	\label{eq:loss_outer}
	\begin{aligned}
	\mathcal{L}_{meta}(\psi'_i; D_i^{val})=\mathcal{L}_{gaze}(\psi'_i; \mathcal{S})+
	\gamma \cdot d(\mathcal{S}, D_i^{val}; \psi'_i),
	\end{aligned}
\end{equation}
where $\mathcal{L}_{gaze}(\psi'_i;\mathcal{S})$ is the standard supervised loss for gaze prediction defined on the labeled source data $\mathcal{S}$. Here we use the angular gaze estimation error in terms of L1 loss for $\mathcal{L}_{gaze}$. Here $d(\mathcal{S}, D_i^{val}; \psi'_i)$ is a measure of the difference between two domains $\mathcal{S}$ and $\mathcal{D}^{val}_i$. Inspired by \cite{long2017deep}, we implement $d(\mathcal{S}, D_i^{val}; \psi'_i)$ using joint maximum mean discrepancy (joint MMD) as:
\begin{equation}
	\label{eq:mmd}
	d(S, D_i^{val}, \psi_i')=
 ||\mathcal{C}_{Z^{1:n}}(S)-\mathcal{C}_{Z^{1:n}}(D_i^{val})||^2,
\end{equation}
here $n$ is meta-batch size. The $\mathcal{C}_{Z^{1:n}}$ is  empirical joint
embedding. For unlabeled person-specific data $D_i^{val}$, it can be estimated as $\mathcal{C}_{Z^{1:n}}(D_i^{val})=\frac{1}{t}\sum_{j=1}^t\otimes_{l=1}^n\phi^l(\mu_j^l)$, where $t$ is number of samples in $D_i^{val}$, and $\mu_j$ is a sample drawn from $D_i^{val}$. It is noted joint MMD does not require labels, but it implicitly depends on $\psi'_i$ since it is defined on the features extracted by the backbone network.
\input{alg1}
\begin{figure}[ht]
	\begin{center}
		\includegraphics[width=.9\linewidth]{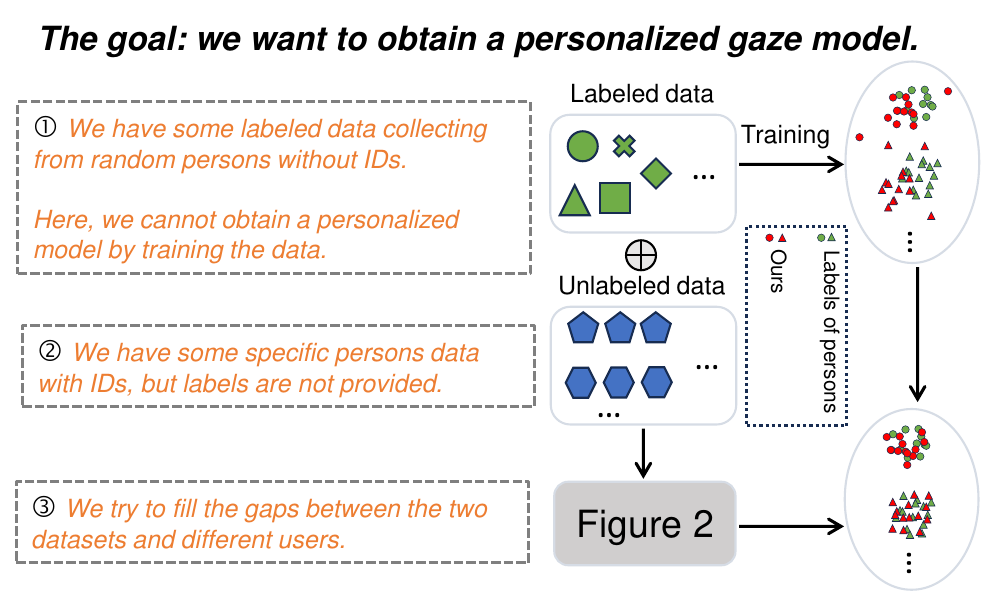}
	\end{center} 
 \vspace{-0.5cm}
	\caption{ { \bf The flow of the proposed method.}  }
 \label{fig:toy}
  \vspace{-0.4cm}
 \end{figure}

The goal of meta-learning is to learn a good initial model $\psi$, so 
that after model update using the support set (Eq.~\ref{eq:inner}) of a 
particular task, the task-adapted model $\psi'_i$ will minimize the loss 
defined in Eq.~\ref{eq:loss_outer} on the corresponding query set across $K$ tasks (persons). The meta-objective is defined as:
\begin{equation}
	\label{eq:meta_obj}
	\min_{\psi}\sum_{i=1}^{K}
	\mathcal{L}_{meta}(\psi'_i; D_i^{val}).
\end{equation}
The initial model $\psi$ is learned by optimizing the meta-objective (Eq.~\ref{eq:meta_obj}) using stochastic gradient descent as:
\begin{equation}
	\label{eq:meta_update}
	\psi \leftarrow \psi - \beta \nabla_{\psi}\sum_{i=1}^N 
	\mathcal{L}_{meta}(\psi'_i; D_i^{val}),
\end{equation}
where $\beta$ represents the meta-learning rate. The meta-objective in Eq.~\ref{eq:meta_obj} 
involves summing over all meta-training tasks. In practice, we sample a mini-batch of tasks (persons) in each iteration during meta-training and sum over these sampled tasks. The entire training procedure is elaborated in Algorithm~\ref{alg:tgkd} and Fig.~\ref{fig:meta}. We also give the flow of the proposed method in Fig.~\ref{fig:toy}.

\noindent{\bf Meta-Testing (Adaptation + Inference).} After meta-training, the initial model parameters $\psi$ have been learned to effectively adapt to a person using only a small number of unlabeled images. During testing (also called meta-testing in our work which consists of adaptation and inference stages), given a new person $\mathcal{D}=(D^{tr}, D^{val})$ with the support set $D^{tr}$ and the query set $D^{val}$. We simply use Eq.~\ref{eq:inner} to obtain the person-specific parameters $\psi'$ via the 
self-supervised loss. The $\psi'$ is adopted for gaze estimation in other face images from this person. Note that we do not need any labels for the adaptation during meta-testing.

%% file: alg1.tex
\begin{algorithm}[t]
	\caption{Training for ELF-UA}
	\label{alg:tgkd}
	\begin{algorithmic}[1]
                \REQUIRE Distribution $p(\mathcal{T})$ over tasks $\mathcal{D}$; labeled source dataset $P_S$; 
		\REQUIRE Learning rates $\alpha$ and $\beta$; meta-batch size $n$
		\STATE Initialize network $\theta$ 
		\WHILE{not done}
		\STATE Sample $K$ unlabeled tasks (i.e. persons) $\mathcal{D}_i \sim p(\mathcal{T})$
		\FORALL{$\mathcal{D}_i$}
                \STATE Construct $\mathcal{D}_i=\{D_i^{tr}, D_i^{val}\}$ 
                \STATE Construct self-supervised loss $\mathcal{L}_{self}(\theta; D_i^{tr})$
                \STATE Obtain the adapted model $\psi'_i$ via 
                Eq.~\ref{eq:inner} on $D_i^{tr}$
                \STATE Construct upper bound of outer loop loss $\mathcal{L}_{meta}(\psi'_i;D_i^{val})$
                \STATE Evaluate meta-objective in Eq.~\ref{eq:loss_outer} on $D_i^{val}$
		\ENDFOR
                \STATE Update model parameters $\psi$ via 
                Eq.~\ref{eq:meta_update}
		\ENDWHILE
	\end{algorithmic}
\end{algorithm}

%% file: sec/4_experiment.tex
\section{Experiments}
In this section, we first introduce the datasets and setup. We then describe our implementation details. We present our experimental results and comparisons. 
 \input{tabs/tab1}

\subsection{Datasets and Setup}\label{sec:dataset}
Since we propose a new problem setting, there have no existing datasets and experimental protocols for this problem. We re-purpose several existing gaze datasets for our problem, including ETH-XGaze~\cite{eth_2020}, Gaze360~\cite{gaze360_2019}, GazeCapture~\cite{2016Krafka} and MPIIGaze~\cite{mpii}. The ETH-XGaze datast ($\mathcal{D}_E$) contains 756,540 images without person IDs. The Gaze360 dataset ($\mathcal{D}_G$) has 84902 images without person IDs. The MPIIGaze dataset ($\mathcal{D}_M$) has 15 subjects, where each subject contains 3000 images. GazeCapture ($\mathcal{D}_{GC}$) is the largest in-the-wild dataset for gaze estimation. It contains data from over 1,450 people consisting of almost 2.5M frames. Since the original GazeCapture dataset only provides gaze labels on a 2D screen, we use the pre-processing pipeline~\cite{park2019few} to attain 3D head pose from GazeCapture. We select 993 persons with over 400 samples per person as the unlabeled person-specific dataset for training. We select 109 persons with over 1000 samples per person are used for testing.

We perform experiments using the following different setups. We use either $\mathcal{D}_{E}$ or $\mathcal{D}_{G}$ as the labeled source data. The $\mathcal{D}_{GC}$ training split (993 persons) is used as the unlabeled person-specific data in all setups. We then test the model on either the $\mathcal{D}_{GC}$ test split (109 persons) or $\mathcal{D}_M$ (15 persons). These different setups will simulate real-world scenarios where data are combined from different sources.

 \subsection{Implementation Details}\label{sec:implementation}
 \noindent{\bf Network Architecture.} We use a ResNet-based~\cite{he2016deep} architecture as the main network $B_{\psi}$ by removing the last fully-connected layer, then add a  two-layer MLP with 256 hidden neurons and output dimension of 128 to the last block of ResNet. The main task outputs the 2-D dimension representing the yaw and pitch angles of the gaze. The auxiliary task (jigsaw puzzle) outputs a $M$-dimension vector representing the predicted permutation. During the training and testing, we only update the parameters for the last ResNet block and the MLP layer.
 
  \input{tabs/tab3}

 \noindent{\bf Training Details.}  We implement our method in PyTorch. All experiments are performed on a single RTX 3090 GPU. We use SGD for the meta-optimization with a fixed learning rate of $\beta=10^{-4}$. The inner optimization is done using 3 gradient steps with an adaptation learning rate of $\alpha=10^{-2}$. During meta-testing, we use the same fixed learning rate and 3 gradient steps as meta-optimization to keep consistent with training. We set the weight of the distance between source and target data to $\gamma$ = 0.1 (Eq.~\ref{eq:loss_outer}). The meta-batch size is set to $n$ = 10 (Eq.~\ref{eq:mmd}) and each task consists of $K = 5$ face images. All images are cropped to the resolution of 224$\times$224. For the auxiliary task, the face image is decomposed to 16 shuffled patches by a regular $4\times4$ grid, and we define $M=31$ classes for the patch permutation prediction task.

\subsection{Results and Comparison}
\label{sec:result}
Since this paper addresses a new problem in gaze estimation, there is no previous work that we can directly compare with. Nevertheless, we define several baseline methods for comparison. We also modify some state-of-the-art domain adaptation approaches for comparison. 

\textbf{Baseline:} This method uses the same ResNet-18~\cite{eth_2020} backbone as ours and is trained on the source data in a standard supervised manner.
\textbf{Ours (w/o adaptation):} This is similar to our approach. The only difference is that it does not perform adaptation at test time.
\textbf{Oracle:} We also consider an oracle method for comparison. This oracle is the same as ours during meta-training. But during meta-testing, it has access to the ground-truth labels of the images in the support set. It uses these labeled images (instead of unlabeled images in our case) to fine-tune the model with a supervised loss for adaptation.
\textbf{PnP-GA:}~\cite{liu2021generalizing} This is a plug-and-play gaze estimation adaptation framework, which has access to 10 pre-trained models with about 100 target samples at test time.
\textbf{DAGEN:}~\cite{guo2020domain} It is a state-of-the-art (SOTA) unsupervised domain adaptation gaze estimation model, which needs to access the 500 target samples for adaptation.
\textbf{UMA:}~\cite{cai2020generalizing} It is an unsupervised domain adaptation using self-training to decrease domain shift, which needs around 100 target samples for adaptation.
\textbf{GVBGD:}~\cite{cui2020gradually} It is a SOTA unsupervised domain adaptation method for classification, which has access to 1000 target samples for adaptation.

Since some of above methods do not address gaze prediction (UMA~\cite{cai2020generalizing} and GVBGD~\cite{cui2020gradually}), we retrain these models with gaze datasets. For a fair comparison, we use ResNet-18 (our backbone) to replace their original networks.

\noindent{\bf Comparison with Baselines.} Table~\ref{tab:1} shows the results of the proposed method and baseline methods.  The baseline (ResNet-18) is trained on the source dataset (ETH-XGaze and Gaze360). ``Ours'' and ``Ours (w/o adaptation'' are trained on the labeled source dataset and unlabeled person-specific datasets (GazeCapture). ``Oracle'' is same as ours during meta-training, but it has access to the labels of the support set during meta-testing. So ``Oracle'' provides an upper bound for our approach. Our proposed method achieves significant improvement over the baseline. It also performs better than ``Ours (w/o adaptation)'' for in all cases. Due to space limitations, we highly refer the readers to the supplementary material for detailed results of each subject (person) on the MPIIGaze dataset.

\noindent{\bf Comparison with SOTA Methods.} Table~\ref{tab:3} shows the quantitative results of different state-of-the-art methods. We compare our method with four supervised gaze estimation methods, including ResNet-50~\cite{eth_2020}, Full-Face~\cite{2017Zhang4}, MANet~\cite{wu2022} and Gaze360~\cite{gaze360_2019}. The results are obtained from either the original papers or running official codes. 
We also compare with some domain adaptation methods, including Gaze360~\cite{gaze360_2019}, DAGEN~\cite{guo2020domain}, GVBGD~\cite{cui2020gradually}, UMA~\cite{cai2020generalizing} and PnP-GA~\cite{liu2021generalizing}. Note that domain adaption methods in Table~\ref{tab:3} require more information that our method. All of them require enough unlabeled data (at least 100) in the target domain and access to the source data during adaptation. They also require heavy computation during the adaptation since they need to run gradient updates for many iterations. So Please refer to Table~\ref{tab:diff} for detailed comparisons of these settings. Nevertheless, the proposed model still achieves competitive performance compared with other SOTA methods. Due to space limitations, many ablation studies are available in the supplementary material.

%% file: tabs/tab1.tex
\begin{table}[htbp]
\caption{ {\bf Comparison results with baseline and oracle methods.} We use either ETH-XGaze ($\mathcal{D}_E$) or Gaze360 ($\mathcal{D}_G)$ as the labeled source data. 
The training split (903 persons) from GazeCapture ($\mathcal{D}_{GC}$) is used as the unlabeled person-specific data. 
Angular gaze error ($^{\circ}$) is used as the evaluation metric.}
 \vspace{-0.5cm}
\label{tab:1}
	\begin{center}
            \small
            \resizebox{1\linewidth}{!}{
		\begin{tabular}{lcccc}
			\toprule[1.2pt]
			 Method & $\mathcal{D}_{E}\rightarrow\mathcal{D}_M$ & $\mathcal{D}_{E}\rightarrow\mathcal{D}_{GC}$ & $\mathcal{D}_{G}\rightarrow\mathcal{D}_M$ & $\mathcal{D}_{G}\rightarrow\mathcal{D}_{GC}$ \\	
			\hline
			\specialrule{0em}{1pt}{1pt}	
			 Baseline (ResNet-18)      & 8.36 & 11.79 & 8.39 & 10.53 \\
			 \textbf{Ours} & \textbf{6.79} \upscore{18.7} & \textbf{7.90} \upscore{33.0} & \textbf{7.38} \upscore{12.0} & \textbf{9.12} \upscore{13.4} \\\hline\specialrule{0em}{1pt}{1pt}
			
			Ours (w/o adaptation)      & 7.16 & 8.63 & 7.89 & 9.26 \\
			 \textbf{Ours}  & \textbf{6.79} \upscore{5.2} & \textbf{7.90} \upscore{8.5} & \textbf{7.38} \upscore{6.5} & \textbf{9.12} \upscore{1.5} \\\hline \specialrule{0em}{1pt}{1pt}
			
			 Oracle   & 6.11 & 7.23 & 6.45 & 7.56 \\
			\bottomrule[1.2pt]
		\end{tabular}}
	\end{center}
 \vspace{-0.5cm}
\end{table}

%% file: tabs/tab3.tex
\begin{table*}[t]
	\renewcommand\arraystretch{1.3}
	\setlength\tabcolsep{12pt}
	\normalsize
	\caption{{\bf Performance comparison with SOTA methods.} Since our problem setting is new, we compare with existing supervised methods and domain adaptation methods. Our method achieves a competitive result. Note that the domain adaptation methods use more information than ours. In particular, they require access to the source domain and enough unlabeled data in the target domain during adaptation. They are also computationally expensive. Table~\ref{tab:diff} provides details for these settings.  We report the angular error ($^\circ$) (lower is better) as the evaluation metric.}
\centering
 \vspace{-0.3cm}
\resizebox{.9\linewidth}{!}{
	\begin{tabular}{p{1.9cm}|p{4.5cm}|c|c|c|c|c|c}	
 
		\toprule[1.2pt]
		Category & Methods & Model &Target samples & {$\mathcal{D}_{E}$ $\rightarrow$ $\mathcal{D}_M$} & {$\mathcal{D}_E$$\rightarrow$$\mathcal{D}_{GC}$}&{$\mathcal{D}_G$$\rightarrow$$\mathcal{D}_M$}& {$\mathcal{D}_G$$\rightarrow$$\mathcal{D}_{GC}$} \\
  
		\hline
		\multirow{6}{*}{\tabincell{c}{Without \\adaptation}}
    & ResNet-50~\cite{eth_2020}&Agnostic &N/A &8.02 &10.5 &8.33&12.9\\
		&Full-Face\cite{2017Zhang4}&Agnostic&N/A&12.35& 14.52 &11.13 & 16.12\\
		&MANet\cite{wu2022}&Agnostic&N/A&7.25 &9.88 &\underline{7.96}&\underline{9.86}\\
		&Gaze360\cite{gaze360_2019}&Agnostic&N/A&\underline{7.23}& \underline{9.02} & 11.36 & 15.86 \\
		\cline{2-8}
		&Baseline\cite{eth_2020}&Agnostic&N/A&8.36 &11.79& 8.39&10.53\\
		&\textbf{Ours (w/o adaptation)}&Agnostic&N/A& \textbf{7.16} & \textbf{8.63} & \textbf{7.89} & \textbf{9.26}\\
		
		\hline
		\hline
		\multirow{6}{*}{\tabincell{c}{With \\ adaptation}}
		&Gaze360\cite{gaze360_2019}&Dataset-specific&$>$100&10.23 &9.82&7.48 &\underline{7.01}\\
		&DAGEN\cite{guo2020domain} &Dataset-specific&$\sim$500 & 7.77 & 8.23 &7.96 &9.35\\
		&GVBGD\cite{cui2020gradually}&Dataset-specific&$\sim$1000 &8.63 &8.10 &7.78 &8.20\\
		&UMA\cite{cai2020generalizing}&Dataset-specific&$\sim$100 &9.96 &9.42 &8.37 &7.54\\
		&PnP-GA\cite{liu2021generalizing} &Dataset-specific&$\sim$100&\underline{7.11} &\underline{7.97} &\textbf{6.58} &\textbf{6.98}\\
		\cline{2-8}
		&\textbf{Ours}&Person-specific&5& \textbf{6.79} & \textbf{7.90} &\underline{7.38} &9.12
		\\
		\toprule[1.2pt]
	\end{tabular}}
	\label{tab:3}
 \vspace{-0.3cm}
\end{table*}

%% file: sec/5_conclusion.tex
\vspace{-0.3cm}
\section{Conclusion}
We have proposed a new problem setting for efficient label-free user adaptation in gaze estimation. Previous work on user-adaptive gaze estimation requires labeled target data to fine-tune the model at test time. Besides, previous work requires large-scale training data containing gaze labels and person IDs. This is unrealistic in real-world scenarios. Our proposed new setting addresses these issues. We propose a novel meta-learning approach for this new problem. Our approach uses some labeled source data without person IDs and some unlabeled person-specific data for training. Our method uses a generalization bound from domain adaptation to define the loss function in meta-learning. This allows us to learn how to adapt to a new user with only a few unlabeled images even when we do not have training data annotated with both gaze labels and person IDs\\
\noindent{\bf Acknowledgements.} This work is supported by the National Natural Science Foundation of China (No. 62372329), in part by Shanghai Rising Star Program (No.21QC1400900), in part by Tongji-Qomolo Autonomous Driving Commercial Vehicle Joint Lab Project, and in part by Xiaomi Young Talents Program.

%% file: sec/6_appendix.tex




\renewcommand{\thesection}{\Alph{section}}
\section*{\Large Supplementary Material }
Due to space limitations in the main paper, we are unable to include all the details there. Here we provide additional (a) detailed performance of each subject (person) on the MPIIGaze dataset, (b) ablation studies.

\section{Experimental Results}
We report the detailed performance of each subject (person) on the MPIIGaze dataset in Table~\ref{tab:2}. Our method successfully adapt to each person and outperforms other alternative methods.

\section{Ablation Studies}

\begin{figure}[htbp]
	\centering
	\subcaptionbox{\label{fig:hypers1}}{\includegraphics[width = 
		.5\linewidth]{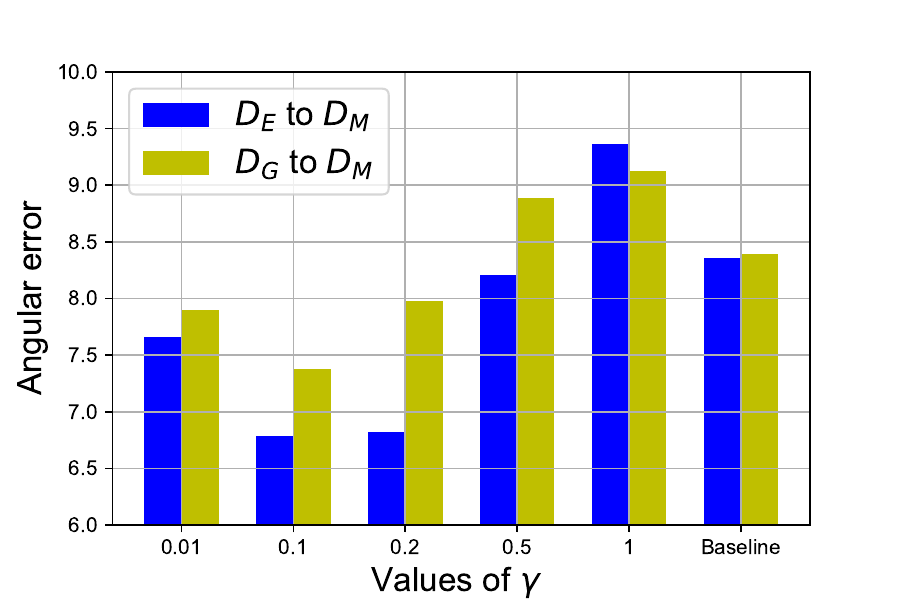}}\hfill
	\subcaptionbox{ \label{fig:hypers2}}{\includegraphics[width = 
		.5\linewidth]{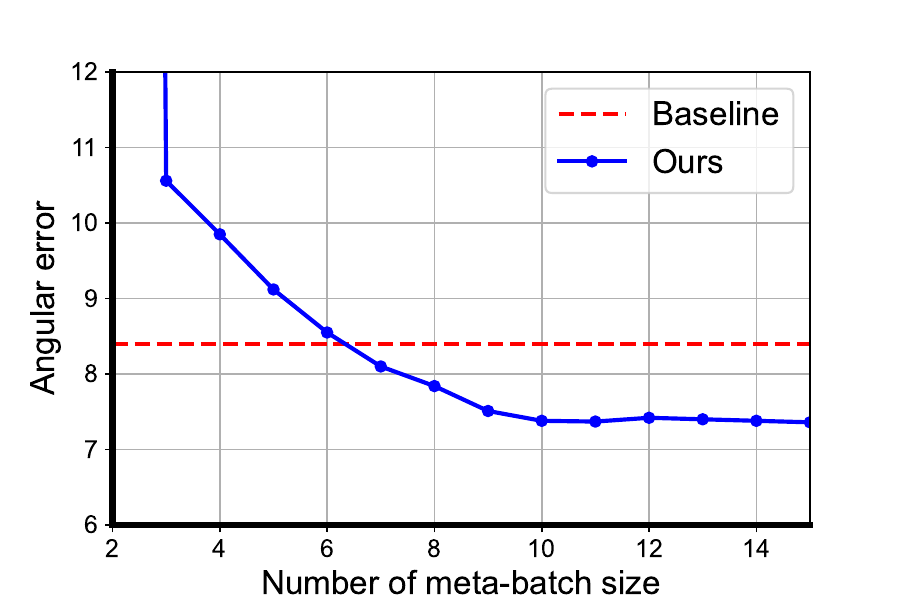}}
	\caption{{\bf Performance varies with different values for balance weight $\gamma$ and meta-batch size}. We 
	show results of $D_E \rightarrow D_M$ and $D_G \rightarrow D_M$ for different values of $\gamma$ (a) and $D_G \rightarrow D_M$ for different numbers of meta-batch size (b).}
 
	\label{fig:hypers}
\end{figure}

\noindent{\bf Hyperparameters.} We evaluate how the performance varies with different balance weight $\gamma$ (Eq.~4) between the labeled source dataset and the unlabeled person-specific dataset. We implement the experiments with different values of $\gamma$, and the results are shown in Fig.~\ref{fig:hypers1}. It shows the proposed model achieves the best results at the value $\gamma=0.1$. We observe that large $\gamma$ tends to produce inferior results. Our hypothesis is that larger $\gamma$ allows the model to bridge the gap between the source dataset (labeled) and unlabeled person-specific datasets, while the model does not focus on learning for gaze direction.

\noindent{\bf Effect of Meta-Batch Size.} To evaluate how the performance with gradually increasing the meta-batch size (number of tasks), we conduct the experiment on the setting of  $\mathcal{D}_G\rightarrow\mathcal{D}_M$. Fig.~\ref{fig:hypers2} provides the detailed results. It indicates that the performance of the proposed model improves when the meta-batch size increases. This is because a larger number of tasks brings diverse data distribution in each iteration which prevents over-fitting.

\noindent{\bf Number of Gradient Updates and K-shot.} We study the impacts of two factors in Algorithm 1 of the main paper: the number of gradient updates and samples (K-shot) for each task in the inner loop. Table~\ref{tab:4} shows the results of models that are trained using various numbers of gradient updates $G= \{1,2,3\}$ and samples $K=\{1,3,5\}$. During training and testing, the number of gradient updates is consistent. Overall, large $K$ and $G$ tend to produce better results. In this work, we only need 5 images (5-shot) to adapt the model.

\input{tabs/tab4}

\input{tabs/tab2}

%% file: tabs/tab4.tex
\begin{table}[htbp]
\centering
\caption{{\bf Evaluation of number of updates G and few-shot (K).} Larger G and K produce better performance. We report the
angular error ($^\circ$) (lower is better) as the evaluation metric.}
\resizebox{0.85\linewidth}{!}{
\begin{tabular}{c|cccc} 
\toprule[1.2pt]
 Methods & $\mathbf{D}_{E}\rightarrow\mathcal{D}_M$ &  $\mathbf{D}_{E}\rightarrow\mathcal{D}_{GC}$ &  $\mathbf{D}_{G}\rightarrow\mathcal{D}_M$ &  $\mathbf{D}_{G}\rightarrow\mathcal{D}_{GC}$ \\ 
 \hline \specialrule{0em}{1pt}{1pt}
  Baseline &8.36 &11.79 &8.39 & 10.53\\
  \hline 
 G=1, K=5  & 7.58 & 8.10 & 8.13 & 9.61\\
 G=2, K=5   & \underline{7.36} & \underline{7.92} & 7.89 & \underline{9.21}  \\
G=3, K=1  & 7.60 & 9.33 & 8.23 & 10.21  \\
G=3, K=3  & 7.43 & 8.03 & \underline{7.66} & 9.78  \\
 G=3, K=5  & \textbf{6.79} & \textbf{7.90} & \textbf{7.38} & \textbf{9.12} \\
\toprule[1.2pt]
\end{tabular}}
\arrayrulecolor{black}
\label{tab:4}
\end{table}

%% file: tabs/tab2.tex
\begin{table*}
  \centering
  \caption{{\bf Per-subject results on the MPIIGaze dataset}. The proposed model is trained with labeled source dataset ETH-XGaze (Gaze360) and unlabeled person-specific dataset GazeCapture. We evaluate each trained model on the 15 subjects from the MPIIGaze dataset. We report the angular error ($^\circ$) (lower is better) as the evaluation metric. }
  \resizebox{1\linewidth}{!}{
  \begin{tabular}{clcccccccccccccccr}
    \toprule[1.2pt]
    Dataset &Method & P00 & P01 & P02 & P03 & P04& P05 & P06 & P07& P08& P09 &P10 & P11 
    & P12 &P13 &P14 &Avg.
    \\
    \midrule
    \multirow{4}{*}{$\mathcal{D}_{E}\rightarrow\mathcal{D}_M$}
     \hspace{3pt}&Baseline (ResNet-18) &6.57 &7.26  &12.81  &6.20  &5.60   
    &7.21  &6.26  &8.93  &10.49  &7.12  &\textbf{5.58}  &12.69 &8.20 & 6.96&13.46 &8.36\\
      \hspace{3pt}&Ours (w/o adaptation) & 5.99 & 6.42 & \textbf{10.72} & 5.88 & 4.00 & 
    6.13 & 5.98 & 7.61 & 8.87& 7.17 & 6.79 & 8.53 &6.71&6.45 &10.12 &7.16\\
    \hspace{3pt}&\textbf{Ours} & \textbf{5.61} &\textbf{6.21} & 10.89  & \textbf{4.68} 
    &\textbf{3.75}&\textbf{5.95} & \textbf{5.60} & 
    \textbf{7.31} & \textbf{8.16} & \textbf{6.61} & 6.29 &\textbf{8.50} & \textbf{6.26} 
    &\textbf{6.15}& \textbf{9.50} &\textbf{6.79}\\ 
    \cline{2-18}
    \hspace{3pt}&Oracle &5.50  &5.51  &8.89  & 3.97 &3.56  &5.52  &4.88  & 6.69 
    & 7.52 &6.02  & 5.48 & 8.01 & 5.86&5.33 &8.89 &6.11 \\ 
    \midrule
    \multirow{4}{*}{$\mathcal{D}_{G}\rightarrow\mathcal{D}_M$}
    \hspace{3pt}&Baseline (ResNet-18) &6.51 &5.95 & 6.68 &11.17 &7.41  &9.62  
    &7.07  &\textbf{7.52}  & 10.24&9.83  &\textbf{7.57}  & 10.16  &\textbf{9.07} &7.09&9.93  &8.39\\
    \hspace{3pt}&Ours (w/o adaptation) & 6.57 & 6.22 & 5.73 & 8.87 & 6.74 & 
    6.86 & 8.05 & 
    8.00 & 8.49& 8.18 & 8.96 & 9.36 &9.65&8.09 &8.52 &7.89\\
    \hspace{3pt}&\textbf{Ours} & \textbf{6.30} & \textbf{5.67} & \textbf{5.63}  & \textbf{8.61} 
    &\textbf{5.93} &\textbf{6.39} & \textbf{7.04} & 
    7.78 & \textbf{7.77} & \textbf{7.57} & 8.54& \textbf{8.65} & 9.38 
    &\textbf{7.06}& 
    \textbf{8.31} &\textbf{7.38}\\ 
     \cline{2-18}
    \hspace{3pt}&Oracle &5.23  &5.31  &4.98  &7.63  &5.55 &6.01 &6.21 &6.36 
    &6.69  & 6.96 & 6.10& 7.23&8.66 &6.68 &7.22 &6.45 \\ 
   \toprule[1.2pt]
  \end{tabular}  }  
  \label{tab:2}
\end{table*}

%% file: main.bbl
\begin{thebibliography}{}

\bibitem[\protect\citeauthoryear{An \bgroup \em et al.\egroup
  }{2021}]{an2021conditional}
Yuexuan An, Hui Xue, Xingyu Zhao, and Lu~Zhang.
\newblock Conditional self-supervised learning for few-shot classification.
\newblock In {\em IJCAI}, 2021.

\bibitem[\protect\citeauthoryear{Ben-David \bgroup \em et al.\egroup
  }{2010}]{ben2010theory}
Shai Ben-David, John Blitzer, Koby Crammer, Alex Kulesza, Fernando Pereira, and
  Jennifer~Wortman Vaughan.
\newblock A theory of learning from different domains.
\newblock {\em Machine learning}, 79:151--175, 2010.

\bibitem[\protect\citeauthoryear{Cai \bgroup \em et al.\egroup
  }{2019}]{cai2019learning}
Ruichu Cai, Zijian Li, Pengfei Wei, Jie Qiao, Kun Zhang, and Zhifeng Hao.
\newblock Learning disentangled semantic representation for domain adaptation.
\newblock In {\em IJCAI}, 2019.

\bibitem[\protect\citeauthoryear{Cai \bgroup \em et al.\egroup
  }{2020}]{cai2020generalizing}
Minjie Cai, Feng Lu, and Yoichi Sato.
\newblock Generalizing hand segmentation in egocentric videos with
  uncertainty-guided model adaptation.
\newblock In {\em IEEE CVPR}, 2020.

\bibitem[\protect\citeauthoryear{Carlucci \bgroup \em et al.\egroup
  }{2019}]{carlucci2019domain}
Fabio~M Carlucci, Antonio D'Innocente, Silvia Bucci, Barbara Caputo, and
  Tatiana Tommasi.
\newblock Domain generalization by solving jigsaw puzzles.
\newblock In {\em IEEE CVPR}, 2019.

\bibitem[\protect\citeauthoryear{Chen \bgroup \em et al.\egroup
  }{2019}]{chen2019closerfewshot}
Wei-Yu Chen, Yen-Cheng Liu, Zsolt Kira, Yu-Chiang Wang, and Jia-Bin Huang.
\newblock A closer look at few-shot classification.
\newblock In {\em ICLR}, 2019.

\bibitem[\protect\citeauthoryear{Chen \bgroup \em et al.\egroup
  }{2020}]{2020Chen}
Meixu Chen, Yize Jin, Todd Goodall, Xiangxu Yu, and Alan~Conrad Bovik.
\newblock Study of 3d virtual reality picture quality.
\newblock {\em IEEE Journal of Selected Topics in Signal Processing},
  14(1):89--102, 2020.

\bibitem[\protect\citeauthoryear{Cheng \bgroup \em et al.\egroup
  }{2020}]{cheng2020coarse}
Yihua Cheng, Shiyao Huang, Fei Wang, Chen Qian, and Feng Lu.
\newblock A coarse-to-fine adaptive network for appearance-based gaze
  estimation.
\newblock In {\em AAAI}, 2020.

\bibitem[\protect\citeauthoryear{Chi \bgroup \em et al.\egroup
  }{2021}]{Chi_2021_CVPR}
Zhixiang Chi, Yang Wang, Yuanhao Yu, and Jin Tang.
\newblock Test-time fast adaptation for dynamic scene deblurring via
  meta-auxiliary learning.
\newblock In {\em IEEE CVPR}, 2021.

\bibitem[\protect\citeauthoryear{Chi \bgroup \em et al.\egroup
  }{2022}]{Chi_2022_CVPR}
Zhixiang Chi, Li~Gu, Huan Liu, Yang Wang, Yuanhao Yu, and Jin Tang.
\newblock Metafscil: A meta-learning approach for few-shot class incremental
  learning.
\newblock In {\em IEEE CVPR}, 2022.

\bibitem[\protect\citeauthoryear{Chi \bgroup \em et al.\egroup
  }{2024}]{chi2024adapting}
Zhixiang Chi, Li~Gu, Tao Zhong, Huan Liu, Yuanhao Yu, Konstantinos~N
  Plataniotis, and Yang Wang.
\newblock Adapting to distribution shift by visual domain prompt generation.
\newblock In {\em ICLR}, 2024.

\bibitem[\protect\citeauthoryear{Cui \bgroup \em et al.\egroup
  }{2020}]{cui2020gradually}
Shuhao Cui, Shuhui Wang, Junbao Zhuo, Chi Su, Qingming Huang, and Qi~Tian.
\newblock Gradually vanishing bridge for adversarial domain adaptation.
\newblock In {\em IEEE CVPR}, 2020.

\bibitem[\protect\citeauthoryear{Finn \bgroup \em et al.\egroup
  }{2017}]{2017_finn_MAML}
Chelsea Finn, Pieter Abbeel, and Sergey Levine.
\newblock Model-agnostic meta-learning for fast adaptation of deep networks.
\newblock In {\em ICML}, 2017.

\bibitem[\protect\citeauthoryear{Funes~Mora \bgroup \em et al.\egroup
  }{2014}]{2014Funes}
Kenneth~Alberto Funes~Mora, Florent Monay, and Jean-Marc Odobez.
\newblock Eyediap: A database for the development and evaluation of gaze
  estimation algorithms from rgb and rgb-d cameras.
\newblock {\em Symposium on Eye Tracking Research and Applications}, 2014.

\bibitem[\protect\citeauthoryear{Gidaris \bgroup \em et al.\egroup
  }{2018}]{gidarisunsupervised}
Spyros Gidaris, Praveer Singh, and Nikos Komodakis.
\newblock Unsupervised representation learning by predicting image rotations.
\newblock In {\em ICLR}, 2018.

\bibitem[\protect\citeauthoryear{Guo \bgroup \em et al.\egroup
  }{2020}]{guo2020domain}
Zidong Guo, Zejian Yuan, Chong Zhang, Wanchao Chi, Yonggen Ling, and Shenghao
  Zhang.
\newblock Domain adaptation gaze estimation by embedding with prediction
  consistency.
\newblock In {\em ACCV}, 2020.

\bibitem[\protect\citeauthoryear{He \bgroup \em et al.\egroup
  }{2016}]{he2016deep}
Kaiming He, Xiangyu Zhang, Shaoqing Ren, and Jian Sun.
\newblock Deep residual learning for image recognition.
\newblock In {\em IEEE CVPR}, 2016.

\bibitem[\protect\citeauthoryear{He \bgroup \em et al.\egroup
  }{2019}]{He_2019_ICCV_Workshops}
Junfeng He, Khoi Pham, Nachiappan Valliappan, Pingmei Xu, Chase Roberts, Dmitry
  Lagun, and Vidhya Navalpakkam.
\newblock On-device few-shot personalization for real-time gaze estimation.
\newblock In {\em IEEE ICCV Workshops}, 2019.

\bibitem[\protect\citeauthoryear{Kellnhofer \bgroup \em et al.\egroup
  }{2019}]{gaze360_2019}
Petr Kellnhofer, Adria Recasens, Simon Stent, Wojciech Matusik, and Antonio
  Torralba.
\newblock Gaze360: Physically unconstrained gaze estimation in the wild.
\newblock In {\em IEEE ICCV}, 2019.

\bibitem[\protect\citeauthoryear{Konrad \bgroup \em et al.\egroup
  }{2020}]{konrad2020gaze}
Robert Konrad, Anastasios Angelopoulos, and Gordon Wetzstein.
\newblock Gaze-contingent ocular parallax rendering for virtual reality.
\newblock {\em ACM Trans. Graph}, 39(2):1--12, 2020.

\bibitem[\protect\citeauthoryear{Krafka \bgroup \em et al.\egroup
  }{2016}]{2016Krafka}
Kyle Krafka, Aditya Khosla, Petr Kellnhofer, Harini Kannan, Suchendra
  Bhandarkar, Wojciech Matusik, and Antonio Torralba.
\newblock Eye tracking for everyone.
\newblock In {\em IEEE CVPR}, 2016.

\bibitem[\protect\citeauthoryear{Lee \bgroup \em et al.\egroup
  }{2019}]{Lee_2019_CVPR}
Kwonjoon Lee, Subhransu Maji, Avinash Ravichandran, and Stefano Soatto.
\newblock Meta-learning with differentiable convex optimization.
\newblock In {\em IEEE CVPR}, 2019.

\bibitem[\protect\citeauthoryear{Liang \bgroup \em et al.\egroup
  }{2020}]{pmlrliang20a}
Jian Liang, Dapeng Hu, and Jiashi Feng.
\newblock Do we really need to access the source data? {S}ource hypothesis
  transfer for unsupervised domain adaptation.
\newblock In {\em ICML}, 2020.

\bibitem[\protect\citeauthoryear{Liang \bgroup \em et al.\egroup
  }{2022}]{liang2022self}
Yudong Liang, Bin Wang, Wangmeng Zuo, Jiaying Liu, and Wenqi Ren.
\newblock Self-supervised learning and adaptation for single image dehazing.
\newblock In {\em IJCAI}, 2022.

\bibitem[\protect\citeauthoryear{Lindén \bgroup \em et al.\egroup
  }{2019}]{2019yu_cvprw}
Erik Lindén, Jonas Sjöstrand, and Alexandre Proutiere.
\newblock Learning to personalize in appearance-based gaze tracking.
\newblock In {\em IEEE CVPR Workshops}, 2019.

\bibitem[\protect\citeauthoryear{Liu \bgroup \em et al.\egroup
  }{2021}]{liu2021generalizing}
Yunfei Liu, Ruicong Liu, Haofei Wang, and Feng Lu.
\newblock Generalizing gaze estimation with outlier-guided collaborative
  adaptation.
\newblock In {\em IEEE ICCV}, 2021.

\bibitem[\protect\citeauthoryear{Liu \bgroup \em et al.\egroup
  }{2024}]{liu2024testtime}
Huan Liu, Julia Qi, Zhenhao Li, Mohammad Hassanpour, Yang Wang, Konstantinos
  Plataniotis, and Yuanhao Yu.
\newblock Test-time personalization with meta prompt for gaze estimation.
\newblock In {\em AAAI}, 2024.

\bibitem[\protect\citeauthoryear{Long \bgroup \em et al.\egroup
  }{2017}]{long2017deep}
Mingsheng Long, Han Zhu, Jianmin Wang, and Michael~I Jordan.
\newblock Deep transfer learning with joint adaptation networks.
\newblock In {\em ICML}, 2017.

\bibitem[\protect\citeauthoryear{Long \bgroup \em et al.\egroup
  }{2018}]{long2018conditional}
Mingsheng Long, Zhangjie Cao, Jianmin Wang, and Michael~I Jordan.
\newblock Conditional adversarial domain adaptation.
\newblock {\em Advances in neural information processing systems}, 31, 2018.

\bibitem[\protect\citeauthoryear{Lu \bgroup \em et al.\egroup
  }{2014}]{lu2014pami}
Feng Lu, Yusuke Sugano, Takahiro Okabe, and Yoichi Sato.
\newblock Adaptive linear regression for appearance-based gaze estimation.
\newblock {\em IEEE Trans. Pattern Anal. Mach. Intell.}, 36(10):2033--2046,
  2014.

\bibitem[\protect\citeauthoryear{Mansour \bgroup \em et al.\egroup
  }{2009}]{mansour2009domain}
Yishay Mansour, Mehryar Mohri, and Afshin Rostamizadeh.
\newblock Domain adaptation: Learning bounds and algorithms.
\newblock In {\em The 22nd Conference on Learning Theory}, 2009.

\bibitem[\protect\citeauthoryear{Munkhdalai and Yu}{2017}]{metanetwork}
Tsendsuren Munkhdalai and Hong Yu.
\newblock Meta neworks.
\newblock In {\em ICML}, 2017.

\bibitem[\protect\citeauthoryear{Nichol \bgroup \em et al.\egroup
  }{2018}]{2018firstorder}
Alex Nichol, Joshua Achiam, and John Schulman.
\newblock On first-order meta-learning algorithms.
\newblock {\em arXiv preprint arxiv:1803.02999}, 2018.

\bibitem[\protect\citeauthoryear{Park \bgroup \em et al.\egroup
  }{2019}]{park2019few}
Seonwook Park, Shalini~De Mello, Pavlo Molchanov, Umar Iqbal, Otmar Hilliges,
  and Jan Kautz.
\newblock Few-shot adaptive gaze estimation.
\newblock In {\em IEEE ICCV}, 2019.

\bibitem[\protect\citeauthoryear{Peng \bgroup \em et al.\egroup
  }{2024}]{peng2024map}
Boyang Peng, Sanqing Qu, Yong Wu, Tianpei Zou, Lianghua He, Alois Knoll, Guang
  Chen, and changjun jiang.
\newblock Map: Mask-pruning for source-free model intellectual property
  protection.
\newblock In {\em IEEE CVPR}, 2024.

\bibitem[\protect\citeauthoryear{Prange \bgroup \em et al.\egroup
  }{2017}]{prange2017speech}
Alexander Prange, Michael Barz, and Daniel Sonntag.
\newblock Speech-based medical decision support in vr using a deep neural
  network.
\newblock In {\em IJCAI}, 2017.

\bibitem[\protect\citeauthoryear{Qu \bgroup \em et al.\egroup
  }{2022}]{qu2022bmd}
Sanqing Qu, Guang Chen, Jing Zhang, Zhijun Li, Wei He, and Dacheng Tao.
\newblock Bmd: A general class-balanced multicentric dynamic prototype strategy
  for source-free domain adaptation.
\newblock In {\em ECCV}, 2022.

\bibitem[\protect\citeauthoryear{Shen \bgroup \em et al.\egroup
  }{2021}]{shen21_aaai}
Chengchao Shen, Xinchao Wang, Youtan Yin, Jie Song, Sihui Luo, and Mingli Song.
\newblock Progressive network grafting for few-shot knowledge distillation.
\newblock In {\em AAAI}, 2021.

\bibitem[\protect\citeauthoryear{Snell \bgroup \em et al.\egroup
  }{2017}]{2017_protonet}
Jake Snell, Kevin Swersky, and Richard Zemel.
\newblock Prototypical networks for few-shot learning.
\newblock In {\em NeurIPS}, 2017.

\bibitem[\protect\citeauthoryear{Sun \bgroup \em et al.\egroup }{2020}]{TTT}
Yu~Sun, Xiaolong Wang, Zhuang Liu, John Miller, Alexei Efros, and Moritz Hardt.
\newblock Test-time training with self-supervision for generalization under
  distribution shifts.
\newblock In {\em ICML}, 2020.

\bibitem[\protect\citeauthoryear{Sung \bgroup \em et al.\egroup
  }{2018}]{2018_reation_net}
Flood Sung, Yongxin Yang, Li~Zhang, Tao Xiang, Philip~H.S. Torr, and Timothy~M.
  Hospedales.
\newblock Learning to compare: Relation network for few-shot learning.
\newblock In {\em IEEE CVPR}, 2018.

\bibitem[\protect\citeauthoryear{Tang \bgroup \em et al.\egroup
  }{2021}]{tang2021self}
Xuwen Tang, Zhu Teng, Baopeng Zhang, and Jianping Fan.
\newblock Self-supervised network evolution for few-shot classification.
\newblock In {\em IJCAI}, 2021.

\bibitem[\protect\citeauthoryear{Vinyals \bgroup \em et al.\egroup
  }{2016}]{2016_matchingnet}
Oriol Vinyals, Charles Blundell, Timothy Lillicrap, Daan Wierstra, et~al.
\newblock Matching networks for one shot learning.
\newblock {\em Advances in Neural Information Processing Systems}, 29, 2016.

\bibitem[\protect\citeauthoryear{Wang \bgroup \em et al.\egroup
  }{2020}]{2019wenguang}
Wenguan Wang, Jianbing Shen, Xingping Dong, Ali Borji, and Ruigang Yang.
\newblock Inferring salient objects from human fixations.
\newblock {\em IEEE Trans. Pattern Anal. Mach. Intell.}, 42(8):1913--1927,
  2020.

\bibitem[\protect\citeauthoryear{Wu \bgroup \em et al.\egroup }{2022}]{wu2022}
Yong Wu, Gongyang Li, Zhi Liu, Mengke Huang, and Yang Wang.
\newblock Gaze estimation via modulation-based adaptive network with auxiliary
  self-learning.
\newblock {\em IEEE Trans. Circuit Syst. Video Technol.}, 32:5510--5520, 2022.

\bibitem[\protect\citeauthoryear{Wu \bgroup \em et al.\egroup
  }{2023}]{wu2022few}
Yong Wu, Shekhor Chanda, Mehrdad Hosseinzadeh, Zhi Liu, and Yang Wang.
\newblock Few-shot learning of compact models via task-specific meta
  distillation.
\newblock In {\em IEEE WACV}, 2023.

\bibitem[\protect\citeauthoryear{Wu \bgroup \em et al.\egroup
  }{2024}]{wu2024testtime}
Yanan Wu, Zhixiang Chi, Yang Wang, Konstantinos~N. Plataniotis, and Songhe
  Feng.
\newblock Test-time domain adaptation by learning domain-aware batch
  normalization.
\newblock In {\em AAAI}, 2024.

\bibitem[\protect\citeauthoryear{Xu \bgroup \em et al.\egroup
  }{2020}]{xu2020joint}
Renjun Xu, Pelen Liu, Yin Zhang, Fang Cai, Jindong Wang, Shuoying Liang, Heting
  Ying, and Jianwei Yin.
\newblock Joint partial optimal transport for open set domain adaptation.
\newblock In {\em IJCAI}, 2020.

\bibitem[\protect\citeauthoryear{Yu \bgroup \em et al.\egroup }{2019}]{2019Yu1}
Yu~Yu, Gang Liu, and Jean-Marc Odobez.
\newblock Improving few-shot user-specific gaze adaptation via gaze redirection
  synthesis.
\newblock In {\em IEEE CVPR}, 2019.

\bibitem[\protect\citeauthoryear{Zhang \bgroup \em et al.\egroup
  }{2017a}]{2017Zhang1}
Xucong Zhang, Yusuke Sugano, and Andreas Bulling.
\newblock Everyday eye contact detection using unsupervised gaze target
  discovery.
\newblock In {\em ACM Symposium on User Interface Software and Technology},
  2017.

\bibitem[\protect\citeauthoryear{Zhang \bgroup \em et al.\egroup
  }{2017b}]{2017Zhang4}
Xucong Zhang, Yusuke Sugano, Mario Fritz, and Andreas Bulling.
\newblock It’s written all over your face: Full-face appearance-based gaze
  estimation.
\newblock In {\em IEEE CVPR Workshops}, 2017.

\bibitem[\protect\citeauthoryear{Zhang \bgroup \em et al.\egroup }{2019}]{mpii}
Xucong Zhang, Yusuke Sugano, Mario Fritz, and Andreas Bulling.
\newblock Mpiigaze: Real-world dataset and deep appearance-based gaze
  estimation.
\newblock {\em IEEE Trans. Pattern Anal. Mach. Intell.}, 41(1):162--175, 2019.

\bibitem[\protect\citeauthoryear{Zhang \bgroup \em et al.\egroup
  }{2020}]{eth_2020}
Xucong Zhang, Seonwook Park, Thabo Beeler, Derek Bradley, Siyu Tang, and Otmar
  Hilliges.
\newblock Eth-xgaze: A large scale dataset for gaze estimation under extreme
  head pose and gaze variation.
\newblock In {\em ECCV}, 2020.

\bibitem[\protect\citeauthoryear{Zhong \bgroup \em et al.\egroup
  }{2022}]{zhong2023metadmoe}
Tao Zhong, Zhixiang Chi, Li~Gu, Yang Wang, Yuanhao Yu, and Jin Tang.
\newblock Meta-dmoe: Adapting to domain shift by meta-distillation from
  mixture-of-experts.
\newblock In {\em NeurIPS}, 2022.

\end{thebibliography}
